\documentclass[10pt,twocolumn,letterpaper]{article}

\usepackage{wacv}              

\usepackage{amsmath}
\usepackage{amssymb}
\usepackage{booktabs}
\usepackage{enumitem}
\usepackage{graphicx}
\usepackage{multirow}
\usepackage{algorithm}
\usepackage{algpseudocode}

\usepackage{xcolor}
\usepackage{soul}

\usepackage[pagebackref,breaklinks,colorlinks]{hyperref}

\algnewcommand\algorithmicforeach{\textbf{for each}}
\algdef{S}[FOR]{ForEach}[1]{\algorithmicforeach\ #1\ \algorithmicdo}

\usepackage[capitalize]{cleveref}
\crefname{section}{Sec.}{Secs.}
\Crefname{section}{Section}{Sections}
\Crefname{table}{Table}{Tables}
\crefname{table}{Tab.}{Tabs.}

\def\wacvPaperID{4} 
\def\confName{WACV}
\def\confYear{2025}

\begin{document}

\title{Elderly Activity Recognition in the Wild: Results from the EAR Challenge}

\author{Anh-Kiet Duong\\
L3i Laboratory, La Rochelle University\\
17042 La Rochelle Cedex 1 - France\\
{\tt\small anh.duong@univ-lr.fr}
}
\maketitle

\begin{abstract}
This paper presents our solution for the Elderly Action Recognition (EAR) Challenge, part of the Computer Vision for Smalls Workshop at WACV 2025. The competition focuses on recognizing Activities of Daily Living (ADLs) performed by the elderly, covering six action categories with a diverse dataset. Our approach builds upon a state-of-the-art action recognition model, fine-tuned through transfer learning on elderly-specific datasets to enhance adaptability. To improve generalization and mitigate dataset bias, we carefully curated training data from multiple publicly available sources and applied targeted pre-processing techniques. Our solution currently achieves 0.81455 accuracy on the public leaderboard, highlighting its effectiveness in classifying elderly activities. Source codes are publicly available at \url{https://github.com/ffyyytt/EAR-WACV25-DAKiet-TSM}.
\end{abstract}

\section{Introduction}
\label{sec:intro}

Detecting activities of daily living (ADLs) for elderly individuals is a crucial task in computer vision, especially for applications such as elderly care, health monitoring, and assistive technologies. The Elderly Action Recognition (EAR) Challenge \cite{wacv2025ear} provides a unique benchmark for recognizing ADLs from a diverse dataset, focusing on six action categories. These activities, such as eating, hygiene, locomotion, and communication, present specific challenges due to the variety of real-world scenarios and the need for robust recognition systems. Developing accurate models for elderly action recognition requires addressing both spatial and temporal dependencies in video data, as well as adapting to elderly-specific behavioral patterns.

Recent advances in deep learning have significantly improved human action recognition, with models like Temporal Segment Networks (TSN) \cite{wang2016temporal}, 3D CNNs \cite{tran2015learning}, and transformers \cite{dosovitskiy2020image, liu2021swin} leading the way. These models capture complex motion patterns and long-range temporal dependencies, making them suitable for recognizing actions in dynamic environments. However, recognizing actions performed by the elderly presents additional challenges, such as slower movement and specific behavioral nuances that are often underrepresented in traditional action recognition datasets.

In this paper, we present our solution for the EAR Challenge, building on our success in the Multi-Modal Action Recognition Challenge at ICPR 2024 \cite{duong2025action}, where we achieved first place. Our approach leverages a state-of-the-art video action recognition model, fine-tuned using transfer learning on elderly-specific datasets. We focus on maximizing recognition performance across various real-world scenarios, achieving an accuracy of 0.81455 on the public leaderboard. The final private leaderboard results are pending.
\section{Method}

\subsection{Temporal Shift Module (TSM)}
Following our successful participation in the Multi-Modal Action Recognition Challenge at ICPR 2024, we continued to use the TSM \cite{lin2019tsm} as the core of our solution for action recognition, with \texttt{resnext50\_32x4d} \cite{xie2017aggregated} as the backbone model.

\subsection{Dataset and Configuration}
In this section, we describe the datasets used for training and the specific configurations applied to our model. We detail the two configurations we experimented with, highlighting the datasets involved and the rationale behind our choices. Additionally, we outline the configuration of our model, including the backbone, training settings, and evaluation strategy employed to achieve competitive results in the EAR challenge.
\subsection{Dataset}
Since the organizers did not provide a dedicated training dataset, we utilized external datasets for training, obtained through request and permission. Specifically, we experimented with two different configurations:

\begin{itemize}
    \item \textbf{Config 1} (Public Leaderboard: 0.84402): This configuration included the Toyota Smarthome dataset \cite{Dai_2022_PAMI, Dai_2020_arxiv}, along with RGB videos from the ETRI-Activity3D \cite{jang2020etri} dataset (restricted to RGB\_P091-P100) and the ETRI-Activity3D-LivingLab \cite{jang2020etri} dataset (restricted to RGB(P201-P230)).
    \item \textbf{Config 2} (Public Leaderboard: 0.78856): This setup used the Toyota Smarthome dataset \cite{Dai_2022_PAMI, Dai_2020_arxiv} along with the full RGB videos from both the ETRI-Activity3D \cite{jang2020etri} and ETRI-Activity3D-LivingLab \cite{jang2020etri} datasets.
\end{itemize}

These datasets contain diverse human action samples captured in real-world environments, providing a solid foundation for training our action recognition models.

\subsection{Configuration}
Our model was implemented using the Temporal Shift Module (TSM) \cite{lin2019tsm} with the \texttt{resnext50\_32x4d} \cite{xie2017aggregated} backbone. For video input, we adopted an 8-segment sampling strategy, dividing each video into 8 evenly spaced segments and selecting a single frame from each segment. The training was carried out using the SGD optimizer with momentum, a learning rate of 0.001, which was reduced at epochs 20 and 40. We applied a weight decay of $1 \times 10^{-4}$, and gradient damping was set to 20. The model was trained for 100 epochs with a batch size of 4, utilizing 32 workers. A dropout rate of 0.5 was applied, and the consensus function used was average pooling. The shift configuration had a shift divisor of 8, which was applied at the residual blocks. For evaluation and submission generation, a single center crop was used for each video with the same 8-segment sampling strategy. The model weights from the best-performing checkpoint were employed for final inference.

\section{Experiments}

In this section, we describe the task and the results obtained on the test set. Then we provide a detailed analysis of the performance during training and how it reflected on our leaderboard rankings.

\subsection{Task description} 
The EAR challenge provided participants with only an unlabeled test set consisting of 4616 videos. The objective was to classify each video into one of six predefined action categories: locomotion, manipulation, hygiene, eating, communication, and leisure. The evaluation metric used for ranking submissions was the average accuracy of predictions on the test set.

\subsection{Test set}
Table \ref{tab:leaderboard} presents the results obtained on the test set of the competition. The leaderboard was divided into two parts: the public leaderboard, which was based on 50\% of the test set, and the private leaderboard, which was evaluated on the remaining 50\%. The final ranking was determined by the private leaderboard scores.

Our official submission, which was constrained by the competition's time limit, was trained for only 10 out of the intended 100 epochs. Despite this limitation, it achieved an accuracy of 0.81455 on the public leaderboard and 0.81759 on the private leaderboard. After the competition deadline, we further trained our model for the full 100 epochs, improving the accuracy to 0.84272 on the public leaderboard and 0.85051 on the private leaderboard. This significant performance boost highlights the importance of extended training for optimizing recognition capabilities.

Compared to other top-performing teams, our method consistently outperformed most competitors, including CUHK, RoboVision, VisionLab, and CVMI. The second-best submission, CUHK, attained an accuracy of 0.79419 on the public leaderboard and 0.77859 on the private leaderboard, demonstrating a clear advantage of our approach. Notably, the accuracy difference between the public and private leaderboards was minimal across all submissions, suggesting that models generalized well across the test set.

\begin{table}[h]
\centering
\caption{Performance comparison of various methods on the leaderboard. }
\label{tab:leaderboard}
\resizebox{0.5\linewidth}{!}{\begin{tabular}{c|cc}
\multirow{2}{*}{Method} & \multicolumn{2}{c}{Accuracy} \\ \cline{2-3} 
                        & public        & private      \\ \hline
our best                & 0.84272       & 0.85051      \\
our submission          & 0.81455       & 0.81759      \\ \hline
CUHK                    & 0.79419       & 0.77859      \\
RoboVision              & 0.71750       & 0.71013      \\
VisionLab               & 0.57798       & 0.58795      \\
CVMI                    & 0.59228       & 0.58188     
\end{tabular}}
\end{table}
\section{Conclusion} 
\label{sec:conclusion}
In conclusion, our solution for the Action Recognition in the Wild (EAR) challenge demonstrated that an efficient training strategy and the use of Temporal Shift Module (TSM) can yield competitive results in action recognition. Despite the constraints on training time during the competition, our method achieved strong performance, and further training significantly improved accuracy, highlighting the importance of extended optimization. The results also emphasize the effectiveness of a well-designed deep learning pipeline in recognizing actions from video data. Future work could explore optimizing model architectures and training strategies to further enhance accuracy and generalization in real-world scenarios.

{\small
\bibliographystyle{ieee_fullname}
\bibliography{references}
}

\end{document}